\documentclass[lettersize,journal]{IEEEtran}
\usepackage{amsmath,amsfonts}
\usepackage{algorithmic}
\usepackage{algorithm}
\usepackage{array}
\usepackage{textcomp}
\usepackage{stfloats}
\usepackage{url}
\usepackage{verbatim}
\usepackage{graphicx}
\usepackage{cite}
\usepackage{graphicx}
\usepackage{subcaption}

\usepackage{color}
\usepackage{colortbl} 
\usepackage{xcolor} 
\usepackage{fancyhdr}

\definecolor{darkblue}{rgb}{0, 0, 0.6}
\definecolor{darkgreen}{rgb}{0, 0.7, 0}
\definecolor{darkred}{rgb}{0.8, 0, 0}
\definecolor{skyblue}{rgb}{0.16, 0.65, 0.87}
\definecolor{orange}{rgb}{0.93, 0.49, 0.19}
\definecolor{green}{rgb}{0.44, 0.68, 0.28}
\definecolor{navy}{rgb}{0.27, 0.45, 0.78}
\usepackage{marvosym}

\newcommand\shline{\specialrule{0.8pt}{0pt}{0pt}}

\newcommand*\colorcheck{%
  \expandafter\newcommand\csname greencheck\endcsname{\textcolor{darkgreen}{\ding{52}}}%
}
\newcommand*\colorcross{%
  \expandafter\newcommand\csname redcross\endcsname{\textcolor{darkred}{\ding{56}}}%
}
\colorcheck
\colorcross

\usepackage{
rotating,
booktabs,
mathrsfs,
url,
etoolbox,
multirow,
hyperref,
manyfoot,
hhline,
pifont,
cleveref}

\hypersetup{
colorlinks=true,
linkcolor=black
}

\begin{document}

\title{TalkPhoto: A Versatile Training-Free Conversational Assistant for Intelligent Image Editing}

\author{Yujie Hu, Zecheng Tang, Xu Jiang, Weiqi Li, and Jian Zhang\textsuperscript{\Letter}\\School of Electronic and Computer Engineering, Peking University}



\maketitle

\begin{abstract}
Thanks to the powerful language comprehension capabilities of Large Language Models (LLMs), existing instruction-based image editing methods have introduced Multimodal Large Language Models (MLLMs) to promote information exchange between instructions and images, ensuring the controllability and flexibility of image editing. However, these frameworks often build a multi-instruction dataset to train the model to handle multiple editing tasks, which is not only time-consuming and labor-intensive but also fails to achieve satisfactory results. In this paper, we present TalkPhoto, a versatile training-free image editing framework that facilitates precise image manipulation through conversational interaction. We instruct the open-source LLM with a specially designed prompt template to analyze user needs after receiving instructions and hierarchically invoke existing advanced editing methods, all without additional training. Moreover, we implement a plug-and-play and efficient invocation of image editing methods, allowing complex and unseen editing tasks to be integrated into the current framework, achieving stable and high-quality editing results. Extensive experiments demonstrate that our method not only provides more accurate invocation with fewer token consumption but also achieves higher editing quality across various image editing tasks. 
\end{abstract}


\section{Introduction}
\IEEEPARstart{I}{mage} editing refers to the process of altering or enhancing images to achieve desired visual effects. It involves a wide range of modifications, from simple adjustments like color correction to more complex changes like facial attribute editing, removing elements, retouching, and restoration. In recent years, the demand for intuitive and efficient image editing tools has skyrocketed. Due to the great progress in text-to-image synthesis \cite{stablediffusion, imagen, dalle2, glide} and multimodal large language models (MLLMs) \cite{llava, blip2, minigpt4, gill, xufakeshield, zhang2025vq, li2025q, zhao2025reasoning, xu2025avatarshield}, instruction-based image editing methods  \cite{prompt2prompt, instructpix2pix, ultraedit, yu2023cross, zhang2022herosnet, yang2025difflle, zhang2024v2a, zhang2025omniguard, peng2022lve, zhang2024gs} are emerging. These methods require large-scale training datasets, considerable computational resources, and specialized expertise, which can hinder their widespread adoption for quick and versatile tasks. Furthermore, most are designed for specific tasks and require edit instructions to be in a fixed format, such as \textit{``Make it a cartoon"}, which limits their flexibility to accommodate a broad range of editing functions. Therefore, there is a strong need for a method that requires no training to achieve complex and flexible editing.

Recent advances in the tool usage research of LLMs \cite{hugginggpt, gorilla} demonstrate excellent performance, positioning them as powerful untrained agents capable of invoking various tools for target tasks. One mainstream approach \cite{visualchatgpt, chameleon} employs function calling in combination with ReAct. Function calling \cite{functioncalling} requires fine-tuning with JSON data for tool invocation, means not all LLMs currently support function calling. Additionally, due to the lack of reasoning capabilities in function calling,  it can only invoke one function at a time and often struggles with handling complex tasks. ReAct \cite{react} supports iterative reasoning and action, enabling multiple function calls. However, it is particularly error-prone when parsing function input arguments, which can lead to situations where the model gets stuck in a loop, repeatedly invoking the same tool.

To address the limitations of the above works, we propose a training-free, multi-functional image editing framework that fulfills the needs of users to freely and conveniently edit real images, termed TalkPhoto. After receiving the user instructions, an LLM serves as the backbone to analyze the user requirements and invoke and combine the corresponding methods under the guidance of a specially designed prompt template. Different from existing methods based on function calling combined with ReAct, we innovatively develop a more simple but efficient prompt template. At the same time, the function input arguments have been simplified, enhancing the stability of invocation. The proposed hierarchical invocation strategy significantly reduces token consumption while facilitating the management of numerous functions. For complex tasks that cannot be handled well by a single method, TalkPhoto can decompose them into multiple subtasks to achieve satisfactory visual results. Additionally, our framework is highly flexible and can continuously add new editing functions without changing any structure, simply by incorporating function implementations and their descriptions. 

Our contributions are summarized as follows:
\begin{itemize}
    \item We propose TalkPhoto, a unified image editing framework without training, which can perform various image editing tasks with just a single text instruction.
    \item We implement plug-and-play and efficient invocation on the open-source LLM, including simple yet effective prompt design, function simplification, and hierarchical invocation. This not only provides the flexibility to add new tools and adapt to new tasks but also achieves more accurate invocation with fewer tokens.
    \item We introduce a new pipeline for instructional image inpainting, combining existing models through LLMs to achieve better object removal results.
    \item Experimental results show that our method achieves more accurate invocation and fewer token consumption, and demonstrates better editing performance.
\end{itemize}

\begin{figure*}[ht]
    \centering
    \includegraphics[width=1\linewidth]{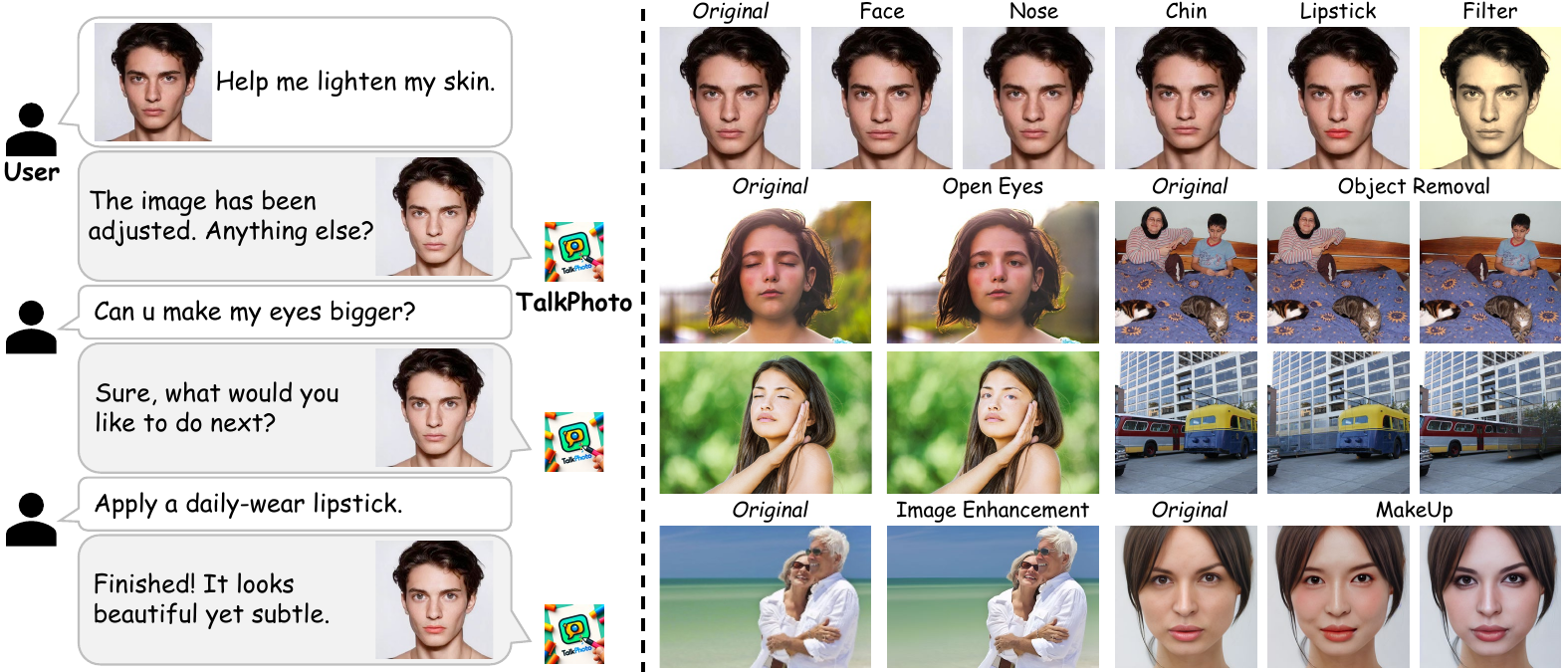}
    \vspace{-16pt}
    \caption{ \textbf{Proposed versatile training-free image editing framework named TalkPhoto.} Left: Only one prompt is needed for users to perform accurate image editing via TalkPhoto. Right: Some supported editing tasks and corresponding results. }
    \vspace{-10pt}
    \label{fig:teaser}
\end{figure*}

\textbf{Organization of This Paper:} In Sec. \ref{sec2}, we briefly review the related works of end-to-end instruction-based image editing and tool-augmented language models. In Sec. \ref{sec3}, we provide a detailed explanation of our proposed method. First, we provide an overview of the framework of TalkPhoto. Then, we detail the plug-and-play and efficient invocation, including prompt design, function simplification, and hierarchical invocation. Finally, we demonstrate the pipeline for instructional image inpainting as an example of how to achieve complex editing tasks. In Sec. \ref{sec4}, we present subjective and objective experimental results, validating the contributions of each innovation. Additionally, we showcase the results of various editing tasks supported by TalkPhoto, along with an interactive demo featuring a user interface. In Sec. \ref{sec5}, we summarize the conclusions of this work.

\section{Related Work}
\label{sec2}

\subsection{Instruction-based image editing}

Editing real images has been widely researched in the field of image processing, with text-guided image editing receiving significant attention due to its flexibility. Some methods\cite{prompt2prompt, imagic, nullinversion} achieve the desired edits in a zero-shot manner. Prompt-to-Prompt \cite{prompt2prompt} modifies the cross-attention maps by comparing additions, deletions, and substitutions of words in the original input caption with the revised caption. Imagic \cite{imagic} fine-tunes the pre-trained text-to-image diffusion model to find a text embedding that represents the input image. It then interpolates linearly between the embedding representing the image and the target text embedding to obtain a semantically meaningful mixture, thereby generating the final editing result. Null Text Inversion \cite{nullinversion} further removes the need for the original caption by using pivotal inversion and null-text optimization. However, detailed editing requires a thorough description of the target image, which can be less user-friendly. Some methods \cite{instructpix2pix, magicbrush, ultraedit} introduce large-scale vision-language image editing datasets to enable direct image editing with simple instructions, allowing users to edit images without requiring elaborate descriptions. InstPix2Pix \cite{instructpix2pix} trains an image editing diffusion model on a paired dataset of text editing instructions and images before/after editing, created by fine-tuned GPT-3 \cite{gpt3} and Prompt-to-Prompt \cite{prompt2prompt} with Stable Diffusion \cite{stablediffusion}. MagicBrush \cite{magicbrush} introduces the first large-scale, manually annotated dataset for instruction-guided real image editing, covering diverse scenarios. UltraEdit \cite{ultraedit} provides the largest instruction-based image editing dataset to be released to the public and proposes a systematic approach to producing massive and high-quality image editing samples automatically. However, these methods typically require end-to-end training and have limited generalization capabilities. Additionally, these instructions are often simple and fixed in format, failing to adapt to diverse user input commands.

\subsection{Tool-augmented language models}
Large language models (LLMs) have made great progress in recent years and have stimulated research in prompt learning and instruction learning. Recent studies have revealed that LLMs can act as controllers to solve complex AI tasks by using tools, with language serving as a generic interface. HuggingGPT \cite{hugginggpt} and Gorilla \cite{gorilla} can select appropriate models or APIs from open-source model libraries such as HuggingFace \cite{huggingface} to execute various tasks. Visual ChatGPT \cite{visualchatgpt} combines ChatGPT \cite{chatgpt} and visual foundation models, enabling ChatGPT to handle complex visual tasks. ToolFormer \cite{toolformer} constructs tool-use augmented data to train language models to use five tools, including a question-and-answer system, a calculator, a search engine, a machine translation system, and a calendar. GPT4Tools \cite{gpt4tools} employs the Low-Rank Adaptation (LoRA) \cite{lora} optimization to facilitate open-source LLMs in solving a range of visual problems. These models either transform models or APIs into language through extensive prompt engineering, making these descriptions distinguishable by the LLM, or fine-tune the language model with a tool-related instruction dataset to improve the accuracy of tool invocation. Recently, many LLM agents \cite{diffagent, cogagent} specifically designed for particular tasks have also emerged. These agents typically gather extensive pairs of user prompts to build self-instruction datasets and then fine-tune the LLM to expand tool usage in specific domains.

\begin{figure*}[ht]
  \centering
  \includegraphics[width=1\linewidth]{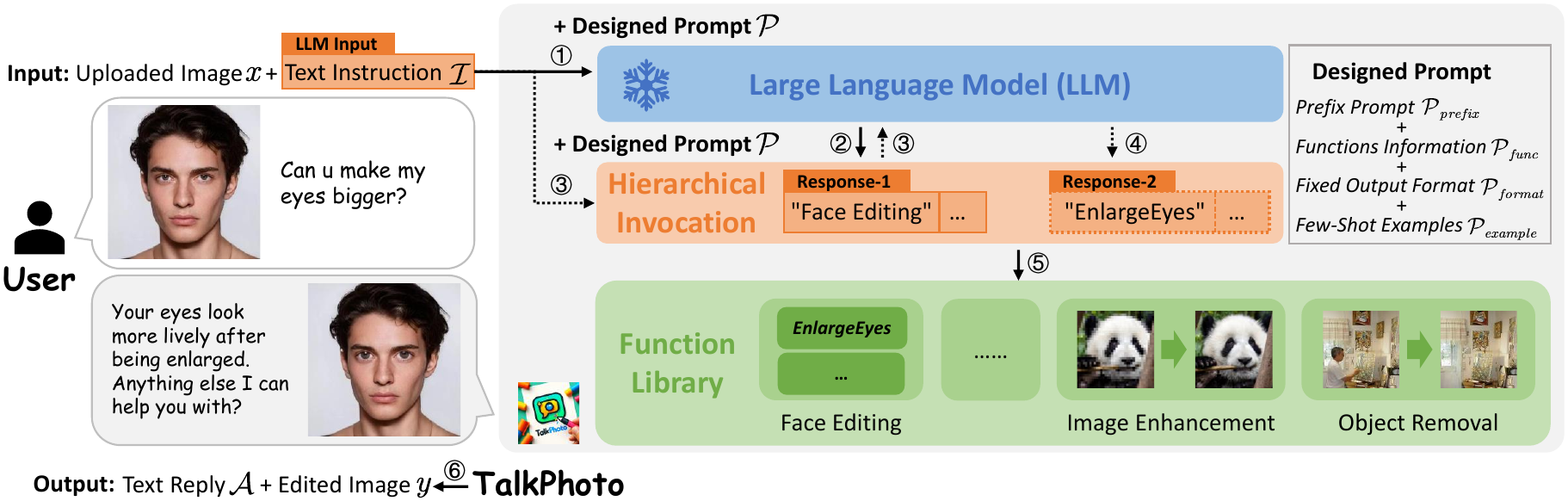}
  \vspace{-16pt}
  \caption{\textbf{Overview of the TalkPhoto framework.} For user input text instructions, we first combine it with a carefully designed prompt and input into the LLM\textsuperscript{\ding{192}} to generate a structured response\textsuperscript{\ding{193}}, detailing the functions to be used along with an analysis and response to the user's needs. If the selected function involves sub-functions or requires additional details, the LLM is consulted a second time to refine the specific functions\textsuperscript{\ding{194}\ding{195}}. The selected function is called from the function library\textsuperscript{\ding{196}} to process the image uploaded by the user. Finally, the edited image and the LLM's response are returned to the user\textsuperscript{\ding{197}}.}
  \vspace{-10pt}
\label{fig-overview} 
\end{figure*}

\section{Method}
\label{sec3}


\subsection{The Framework of TalkPhoto}
The overall framework of TalkPhoto is shown in Fig. \ref{fig-overview}. It consists of four modules: (1) \textit{User Interface}, for receiving user-uploaded images and text instructions, and for displaying the edited images and replies; (2) \textit{LLM}, for responding to text input, analyzing the needs, outputting the functions to be invoked, and returning replies to the user, represented by $LLM$; (3) \textit{Hierarchical Invocation Module}, for engaging in multiple times of interaction with the LLM to determine specific functions, represented by $M$; (4) \textit{Function Library}, for processing images, represented by $F$. The function library consists of various main functions $: \left\{F_1, F_2, \dots, F_N\right\}$, and some main functions are composed of a set of sub-functions $F_n=\left\{f_1, f_2, \dots, f_M\right\}$.

First, the user uploads an image $x$ and provides a text instruction $\mathcal{I}$. The text instruction is then combined with a designed prompt $\mathcal{P}$ and input into the LLM to obtain a fixed-format response $\mathcal{R}$, which includes the functions to be used $\mathcal{R}_{f}$ and the analysis and reply to the user's needs $\mathcal{R}_{a}$. Next, the hierarchical invocation module determines whether the selected function contains sub-functions. If it does, the LLM is invoked twice to further determine the specific functions. The final chosen function is then called from the function library, and once the processing is complete, the edited image $y$ and the LLM's reply $\mathcal{A}$ are returned to the user. The formal definition of TalkPhoto is as follows:
\begin{equation}
\label{eq-talkphoto}
    \left\{y, \mathcal{A}\right\} = \operatorname{TalkPhoto}(x, \mathcal{I}; LLM, M, F, \mathcal{P}).
\end{equation}

\begin{figure}[t]
  \centering
  \includegraphics[width=1\linewidth]{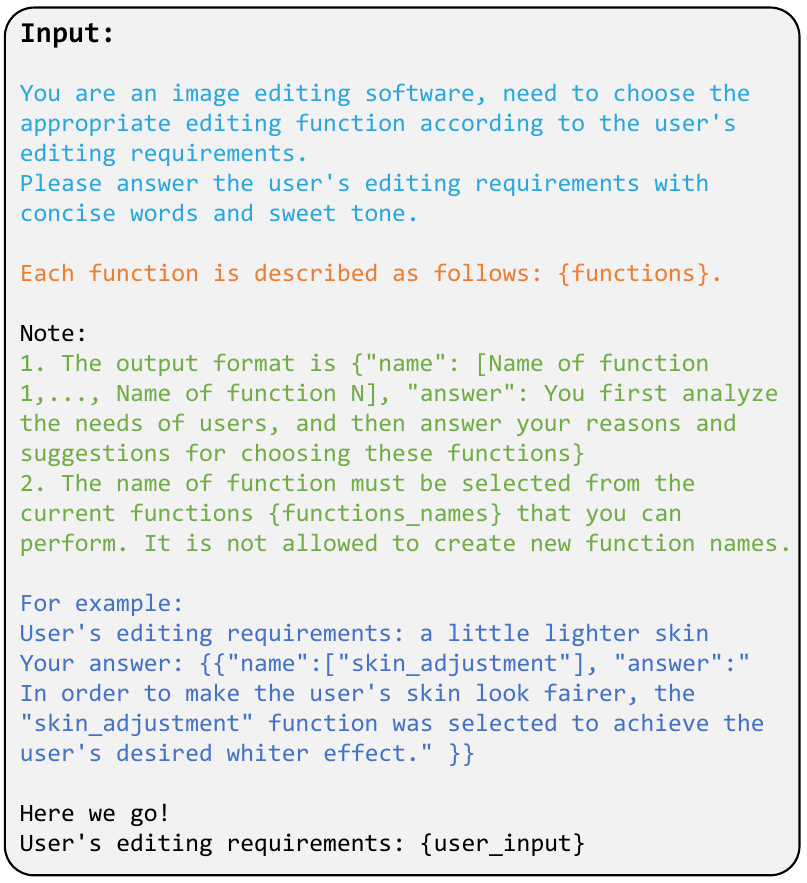}
  \vspace{-16pt}
  \caption{\textbf{The details of the prompt template.} It consists of prefix prompt, functions information, fixed output format, and few-shot examples in turn.}
  \vspace{-10pt}
\label{fig-prompt} 
\end{figure}

\subsection{Plug-and-play and efficient invocation}

\textbf{Prompt design.} The details of the prompt design in TalkPhoto are shown in Fig. \ref{fig-prompt}. Our prompt is divided into four parts: (1) \textcolor{skyblue}{\textit{Prefix Prompt}}, which is mainly to set the role of LLM and tell it what it needs to do, represented by $\mathcal{P}_{pre}$; (2) \textcolor{orange}{\textit{Functions Information}}, including the name and description of the functions contained in the Function Library, represented by $\mathcal{P}_{fun}$; (3) \textcolor{green}{\textit{Fixed Output Format}}, which specifies the output format required from the LLM for easier parsing and subsequent function invocation, represented by $\mathcal{P}_{fmt}$; (4) \textcolor{navy}{\textit{Few-Shot Examples}}, which help the LLM understand and complete the task by providing a few examples, further reinforcing the output format, represented by $\mathcal{P}_{exp}$. Thus, the entire prompt can be represented as: 
\begin{equation}
\label{eq-prompt}
   \mathcal{P}=\operatorname{Concat}(\mathcal{P}_{pre}, \mathcal{P}_{fun}, \mathcal{P}_{fmt}, \mathcal{P}_{exp}).
\end{equation}

Referring to function calling \cite{functioncalling}, we include function names and descriptions in the prompt and ensure that the LLM's output can be parsed by providing few-shot examples and explicitly specifying the output format. More importantly, this way offers plug-and-play flexibility in adding new tools and adapting to new tasks. Inspired by ReAct \cite{react}, we add the sentence, ``\textit{You first analyze the needs of users, and then answer your reasons and suggestions for choosing these functions,}" to our prompt, achieving a reasoning effect. By specifying ``\textit{[Name of function 1,..., Name of function N],}" we also enable the invocation of multiple functions at a time. 

\begin{algorithm}[tb]
\caption{\texttt{Hierarchical invocation} strategy}
\label{algo-invoc}
\textbf{Input}: The large language model $LLM$, the function library $F$, the user text instruction $\mathcal{I}$, the main prompt containing descriptions of main functions $\mathcal{P}_{main}$, the sub-prompt containing descriptions of sub-functions $\left\{\mathcal{P}_{sub}^1, \dots, \mathcal{P}_{sub}^K\right\}$\\
\textbf{Output}: FunctionList
    \begin{algorithmic}[1] 
    \STATE \textcolor{blue}{$\triangleright$ The main invocation}
    \STATE $\mathcal{R}_{f}, \mathcal{R}_{a} \leftarrow LLM(\mathcal{I}, \mathcal{P}_{main})$
    \FOR{$i = 0, \dots, \operatorname{len}(\mathcal{R}_{f}$)}
    \STATE $F_i \leftarrow \operatorname{match}(F, \mathcal{R}_{f}^i)$\\
    \IF{$F_i$ has sub-functions}
    \STATE \textcolor{blue}{$\triangleright$ The sub invocation}
    \STATE $\mathcal{R}_{subf}, \mathcal{R}_{suba}\leftarrow LLM(\mathcal{I}, \mathcal{P}_{sub}^i)$
    \FOR{$j = 0, \dots, \operatorname{len}(\mathcal{R}_{subf}$)}
    \STATE $f \leftarrow \operatorname{match}(F_i, \mathcal{R}_{subf}^{j})$
    \STATE append $f$ to FunctionList
    \ENDFOR
    \ENDIF
    \ENDFOR
    \STATE \textbf{return} FunctionList
    \end{algorithmic}
\end{algorithm}

\textbf{Function simplification.} We observe that LLMs tend to perform better with tasks that involve making selections within a defined range. Therefore, we simplify some functions. For example, in the task of facial whitening, the degree of whitening is usually adjustable. However, spending a lot of text explaining to the LLM the range of possible values, and how adjusting the values will darken or lighten the skin tone, often yields unsatisfactory results. We ultimately decided to split whitening and darkening into two distinct functions and set a degree value with a clear distinction for the input argument. This way, the LLM only needs to determine which function to use and adjust the degree by repeatedly applying the same function. Similarly, with lipstick coloring tasks, it is challenging for people to accurately describe the desired lipstick color, often using subjective expressions like \textit{``a color that brightens the complexion."} Directly parsing the arguments for lipstick coloring might result in varied responses like \textit{``brightening"} or \textit{``lipstick."} We opt to turn common color options into sub-functions, such as \textit{``Pure Red"} or \textit{``Burnt Tomato"}, allowing the LLM to choose the color that best matches the description, thereby addressing the issue of unstable parsing. 

\textbf{Hierarchical Invocation.} A common issue with tool-augmented language models is the maximum token length limit. Users often perform multiple edits on a single image, all within the same conversation round. To maintain context throughout the conversation, tokens accumulate continuously. The function library is also updated regularly. Passing all function descriptions to the LLM increases unnecessary token consumption and reduces the number of edits a user can make. To address these, we propose a hierarchical invocation strategy. The main idea is to group similar functions, including previously mentioned function options, into a main function and manage them through the LLM. An overview of our strategy is presented in Algo.~\ref{algo-invoc}. When a user instruction is received, the LLM first parses the main function from the initial response of the main conversation. If this main function contains sub-functions, a second LLM call is made to locate the specific sub-functions, which starts a new conversation. Compared to not using a hierarchical strategy, passing the names and descriptions of the main functions and sub-functions into two separate conversations can significantly improve the accuracy of function invocation and reduce token consumption. 

\begin{figure}[t]
  \centering
  \includegraphics[width=1\linewidth]{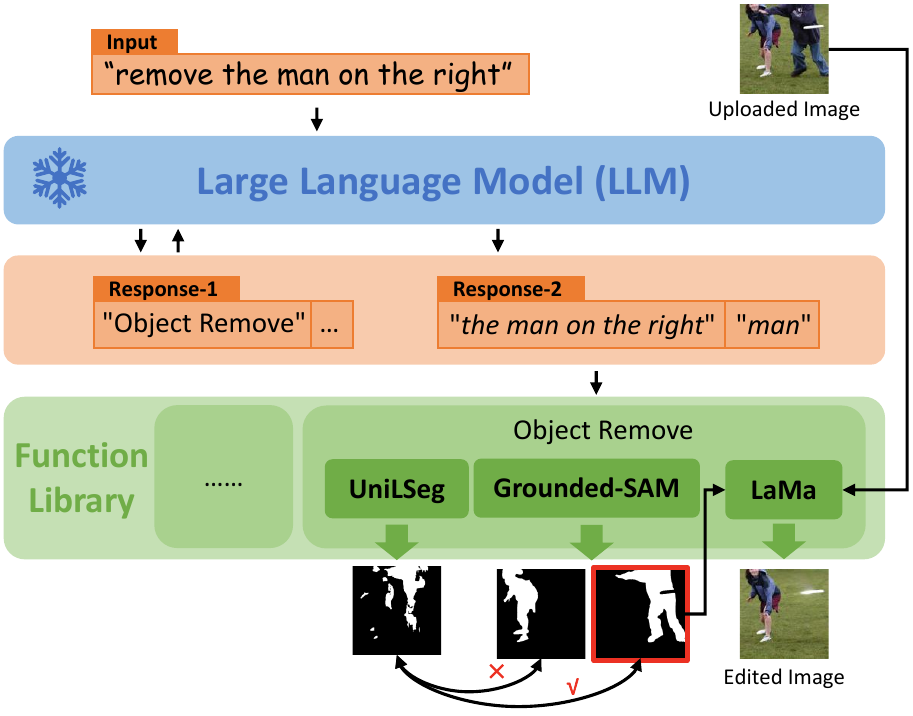}
  \vspace{-16pt}
  \caption{\textbf{Our pipeline for instructional image inpainting.} The LLM responds twice in total. The first response determines the invocation of the object removal function. The second response outputs the category and detailed description of the removed object, which are then input into two segmentation models. The results are matched to obtain the optimal segmentation mask, which is subsequently used for inpainting to generate the final result.}
  \vspace{-10pt}
\label{fig-remove} 
\end{figure}

\subsection{Instructional image inpainting}
\begin{figure}[t]
  \centering
  \includegraphics[width=1\linewidth]{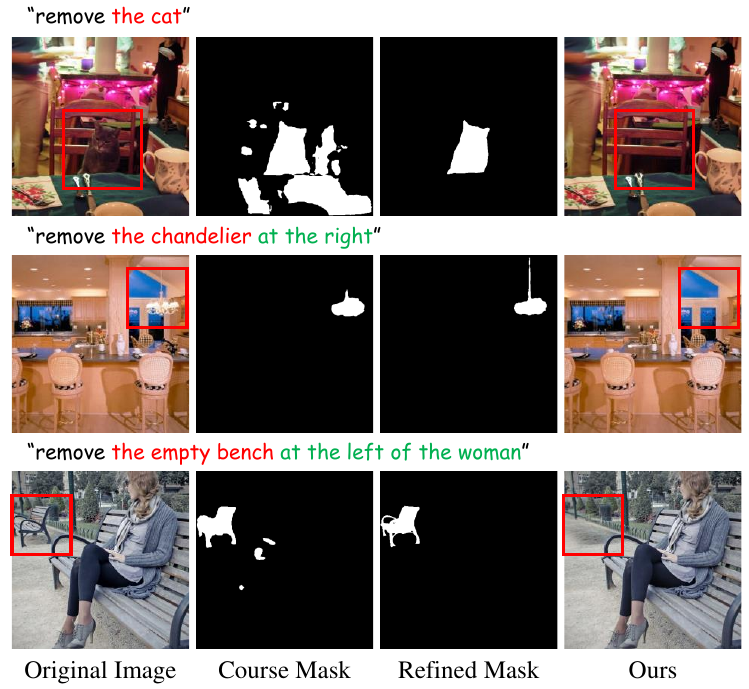}
  \vspace{-16pt}
  \caption{\textbf{Intermediate results of object removal using our method.} The first column is the original image, and the second column is the segmentation mask detected by the pre-trained description-based segmentation model~\cite{unilseg}. The third column shows the refined mask for inpainting. The fourth column is our final result obtained using the pre-trained inpainting model~\cite{lama}.}
  \vspace{-10pt}
\label{fig-remove-mask} 
\end{figure}

Considering that existing methods like referring expression segmentation and binary mask-based inpainting have already achieved good results, we propose a two-stage pipeline for instructional image inpainting \cite{instinpaint}: (1) generating a segmentation mask for the object based on the textual instructions, and (2) using the original image and the mask as input for inpainting. Both of these steps can be directly implemented using existing advanced models. As shown in Fig. \ref{fig-remove}, after the LLM receives the user's instruction to remove a specific object, it first determines to invoke the \textit{``Object Removal"} function. The LLM then provides a second response, analyzing and replying with the category and detailed description of the object to be removed. Next, it calls the category-based segmentation model \cite{unilseg} and description-based segmentation model \cite{groundedsam} separately. Using the coarse mask obtained from the description-based segmentation model, we match it with the mask list generated by the category-based segmentation model, selecting the target with the largest overlapping area as the refined result. The user’s uploaded image and the final refined mask are input into the inpainting model \cite{lama}, and the processed result is the final edited image returned to the user. The intermediate results are demonstrated in Fig. \ref{fig-remove-mask}. The coarse mask obtained by the description-based segmentation model does contain the specified object, but it is not completely accurate. After combining the category-based segmentation model, the final refined mask has a finer outline.

\section{Experiments}
\label{sec4}

\subsection{Implementation details}

\textbf{Supported Functions.} We select the state-of-the-art open-source model, Qwen2-72B-Instruct \cite{qwen2}, as the LLM for TalkPhoto. The following functions are supported:

\textit{- Face Beautification and Shaping, Lipstick Coloring, and Photo Filters}: All of these are implemented by us using the OpenCV library \cite{opencv}. Each operation has predefined parameters, such as the degree of skin whitening in one step, nine lipstick shades, and five filter definitions.

\textit{- One-Click Makeup}: This involves predefined reference images of eleven different makeup styles. We use a makeup transfer method \cite{stablemakeup} to automatically transfer the makeup from the reference image to the input image.
  
\textit{- Image Enhancement}: We use a blind face restoration method \cite{gfpgan} to enhance the face region and a super-resolution method \cite{realesrgan} to enhance the other regions of the image.

\textit{- ID Photo Generation}: Four reference ID photo templates are predefined. We use a face fusion algorithm \cite{facefusion} to generate a new image that resembles the input image's face and has the template's clothing features.
  
\textit{- Open Eyes}: The function first generates a StyleGAN2 \cite{stylegan2} latent vector of the input image, then uses e4e \cite{e4e} to generate an open-eyed face by moving along a predefined latent vector direction corresponding to the \textit{``open eyes"} attribute. The eye region is segmented \cite{farl} (with mask smoothing) and then pasted onto the original image.
  
\textit{- Object Retention and Removal}: This function operates similarly to the object removal function, but instead of performing inpainting, it directly segments out the region of the original image corresponding to the segmentation mask.

\begin{figure}[t]
  \centering
  \includegraphics[width=\linewidth]{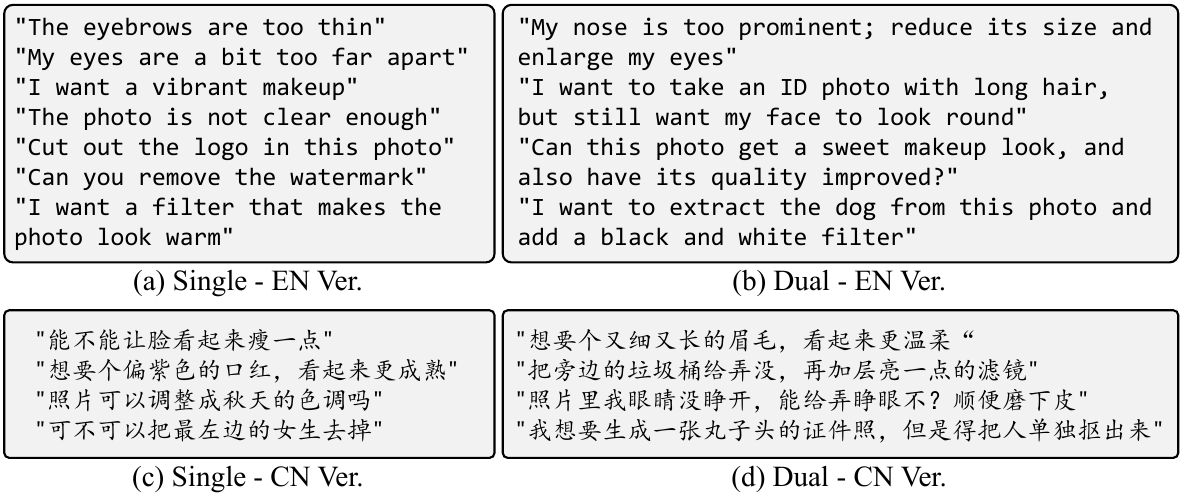}
  \vspace{-16pt}
  \caption{Example instructions of the single-task in English (a) and Chinese (c), and dual-task in English (b) and Chinese(d).}
  \vspace{-10pt}
\label{fig-dataset} 
\end{figure}

\textbf{Datasets.} In order to uniformly evaluate the tool usage frameworks of existing LLMs and TalkPhoto's capability in function invocation, we construct an instruction dataset for testing. The dataset includes single-task and dual-task instructions, each with Chinese and English versions, with 300 examples for each task in each language. Single-task instructions refer to those where fulfilling the user's text instruction requires executing only one function, while dual-task instructions require executing two functions to meet the user's requirements. Example samples are shown in Fig. \ref{fig-dataset}. To assess the performance of TalkPhoto and other instruction-based image editing frameworks in object removal, we select the GQA-Inpaint dataset \cite{instinpaint}, filtering out 101 unique source-target-prompt pairs to ensure that the provided prompts are reasonable and that the mentioned objects are unique. The basic format of the instruction is ``\textit{remove} xxx".

\textbf{Evaluation Metrics.}

\begin{table*}[t]
\centering
\caption{Comparison of Invocation Accuracy and Token Usage across different tasks, methods, and datasets. Function calling only supports single-function calls. ReAct enables multi-function calls, but the strict requirements for parsing function names and arguments lead to a higher error rate. Our method achieves the highest function invocation accuracy and the lowest token consumption.}
\renewcommand{\arraystretch}{1.2}
\resizebox{1.\textwidth}{!}{
\begin{tabular}{c|c|c c|c c|c c}
\shline
\multirow{2}{*}{\textbf{Task}} & \multirow{2}{*}{\textbf{Dataset}} & \multicolumn{2}{c|}{\textbf{Function Calling}} & \multicolumn{2}{c|}{\textbf{ReAct}} & \multicolumn{2}{c}{\textbf{Ours}} \\ \cline{3-8}
&  & Invoc. Acc. & Token Usage & Invoc. Acc. & Token Usage & Invoc. Acc. & Token Usage \\ \hline \hline
\multirow{2}{*}{\textbf{EN Ver.}} & Single & 83.3 & 2879.0 & 56.0 & 6020.2 & \cellcolor[HTML]{EFEFEF}90.0 & \cellcolor[HTML]{EFEFEF}1271.7 \\ 
 & Dual & / & 2888.7 & 46.7 & 6063.3 & \cellcolor[HTML]{EFEFEF}87.3 & \cellcolor[HTML]{EFEFEF}1300.7 \\ \cline{1-8} 
\multirow{2}{*}{\textbf{CN Ver.}} & Single & 87.7 & 2913.2 & 60.0 & 6091.8 & \cellcolor[HTML]{EFEFEF}89.0 & \cellcolor[HTML]{EFEFEF}1299.1 \\ 
 & Dual & / & 2930.6 & 54.7 & 6225.2 & \cellcolor[HTML]{EFEFEF}89.3 & \cellcolor[HTML]{EFEFEF}1377.9 \\ 
\shline
\end{tabular}}
\vspace{-14pt}
\label{tab-func-invoc}
\end{table*}

\begin{figure}[t]
  \centering
  \includegraphics[width=1\linewidth]{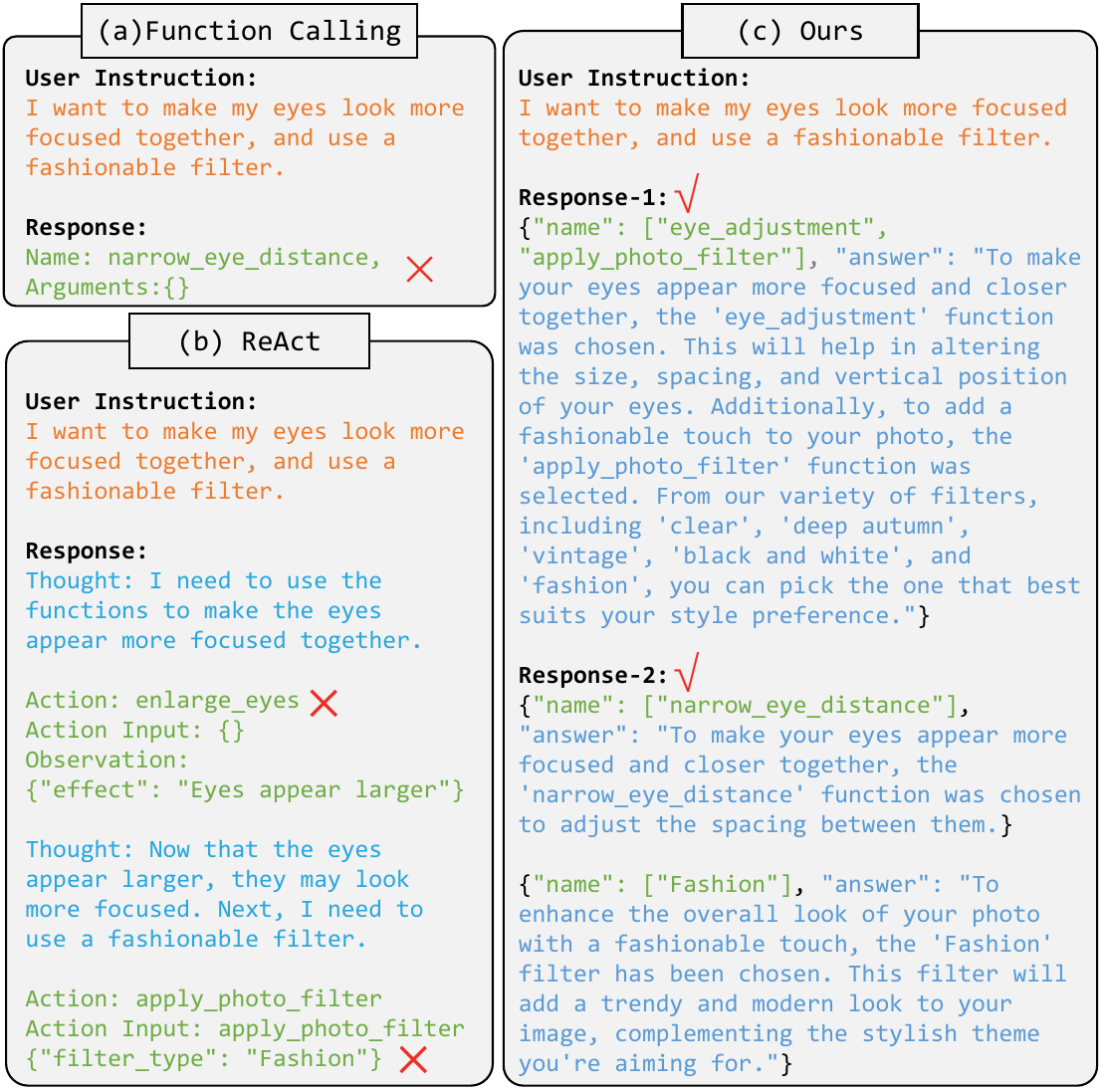}
  \vspace{-16pt}
  \caption{\textbf{Comparison of different function invocations under the same user instruction.} (a) Function calling can only handle single-function calls. (b) ReAct calls are often inaccurate and result in parsing errors. (c) Our method first identifies the main function and then breaks it down into sub-functions, enabling accurate multi-function calls.}
  \vspace{-14pt}
\label{fig-vs-prompt} 
\end{figure}

\subsection{Comparison with the state-of-the-art}

\textbf{Function invocation.} We compare the performance of TalkPhoto with function calling and ReAct on our proposed test instruction dataset. Since Qwen2 supports function calling, we can directly pass in all of our function names and descriptions. ReAct is tested by adding a ReAct prompt template based on the previous function calling. These two methods are currently the primary approaches used in LLMs for tool usage frameworks. As shown in Tab. \ref{tab-func-invoc}, Function calling lacks reasoning capabilities and only supports single-function calls. Adding ReAct enables multi-function calls, but the strict requirements for parsing function names and arguments lead to a higher error rate. With the help of the carefully designed prompt and the hierarchical strategy, our method significantly improves function invocation accuracy and greatly reduces token consumption. As shown in Fig. \ref{fig-vs-prompt}, with the same user instruction, function calling (a) only calls one function, and the called function is wrong. ReAct (b) needs to first determine whether a tool is required; if so, it proceeds with \textit{Action-Action Input-Observation}. It is often error-prone when parsing \textit{Action} and \textit{Action Input}. Our method (c) passing the names and descriptions of the main functions and sub-functions into two separate conversations can significantly improve the accuracy of function invocation. Response-1 and Response-2 are the responses of the main conversation and the second call. 

\begin{figure}[t]
  \centering
  \includegraphics[width=1\linewidth]{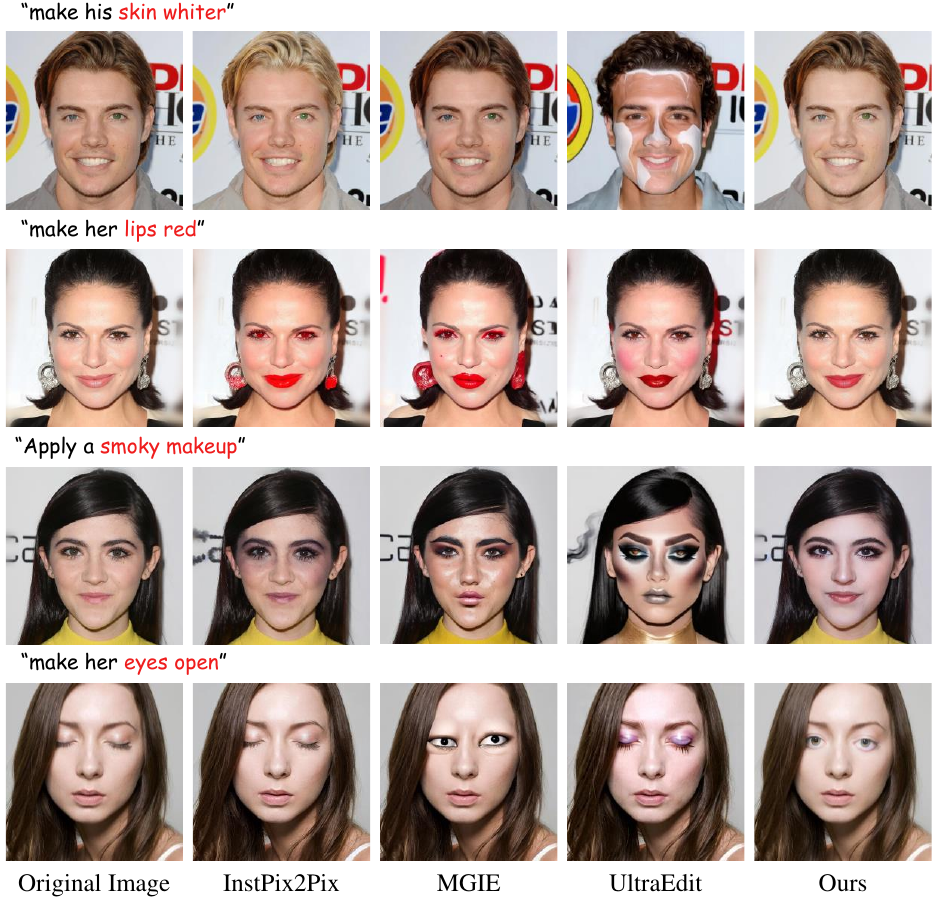}
  \vspace{-16pt} 
  \caption{Qualitative comparison on various editing tasks. End-to-end image editing frameworks \cite{instructpix2pix, mgie, ultraedit} do not align well with the instructions and lack high image quality. Our method achieves better subjective results, especially local region editing.}
  \vspace{-10pt}
\label{fig-vs-multifun}
\end{figure}

\textbf{Instruction-based image editing.} We present the results of some functions and compare them with end-to-end image editing frameworks, including InstPix2Pix  \cite{instructpix2pix}, MGIE \cite{mgie}, and UltraEdit \cite{ultraedit}. Since these frameworks require specific instruction formats during training and inference, we use their required formats to ensure a fair comparison. In practice, TalkPhoto supports flexible instructions and is highly robust to natural language from users. As shown in Fig. \ref{fig-vs-multifun}, the results generated by end-to-end image editing frameworks often do not align well with the instructions and lack high image quality. Our method integrates traditional approaches with advanced deep learning models, achieving better subjective results in various tasks, especially local region editing.

\begin{table*}[t]
\caption{Abalation studies on the core design of function invocation in TalkPhoto. ``Reasoning": whether to include the phrase \textit{``You first analyze the needs of users, and then answer your reasons and suggestions for choosing these functions"} in the prompt template. ``Hierarchy": whether to use a hierarchical strategy for function invocation. ``Examples": how many examples in the prompt template.}
\renewcommand{\arraystretch}{1.2}
\setlength{\tabcolsep}{5mm}
\resizebox{1\linewidth}{!}{
\begin{tabular}{l|ccccccc}
\shline
Config. & (a) & (b) & (c) & (d) & (e) & (f) & (g)\\ \hline \hline
Reasoning & \redcross & \redcross & \greencheck & \greencheck & \greencheck & \cellcolor[HTML]{EFEFEF}\greencheck & \greencheck \\
Hierarchy  & \redcross & \greencheck & \greencheck & \greencheck & \greencheck & \cellcolor[HTML]{EFEFEF}\greencheck & \greencheck \\
Examples  & 0 & 0 & 0 & 1 & 2 & \cellcolor[HTML]{EFEFEF}3 & 4\\ \hline
Invoc. Acc.  & 81.3 & 82.7 & 84.3 & 87.7 & 89.3 & \cellcolor[HTML]{EFEFEF}90.0 & 89.7 \\
Token Usage  & 1775.4 & 927.1 & 952.9 & 1060.6 & 1155.3 & \cellcolor[HTML]{EFEFEF}1271.7 & 1403.2\\ 
\shline
\end{tabular}}
\vspace{-10pt}
\label{tab-aba}
\end{table*}

\begin{figure}[t]
  \centering
  \includegraphics[width=\linewidth]{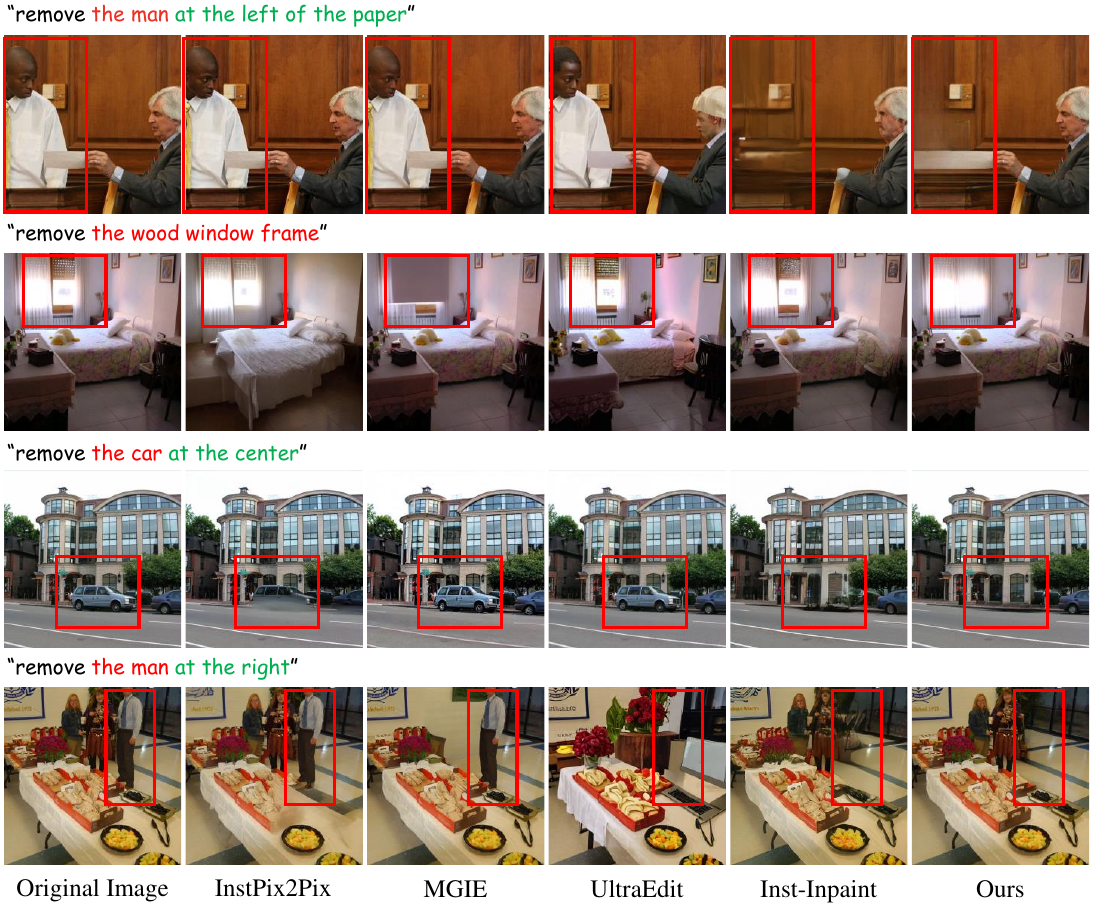}
  \vspace{-16pt}
  \caption{Qualitative comparison on GQA-Inpaint dataset. Our method demonstrates superior editing capabilities compared to existing instruction-based image editing methods. InstPix2Pix \cite{instructpix2pix}, MGIE \cite{mgie}, UltraEdit \cite{ultraedit}, and Inst-Inpaint \cite{instinpaint} are difficult to completely and naturally eliminate the specified object, but our method can, without any visible effect on other areas.}
  \vspace{-10pt}
\label{fig-vs-remove} 
\end{figure}

\begin{table}[h]
\caption{The comparisons with other instruction-based image editing methods on the object removal. The best and the second-best scores are highlighted in \textbf{bold} and \underline{underline} respectively.}
\centering
\renewcommand{\arraystretch}{1.2}
\resizebox{1.0\linewidth}{!}{
\begin{tabular}{l|c|c|c|c}
\shline
Method  & PSNR (dB)$\uparrow$ & SSIM$\uparrow$ & LPIPS $\downarrow$  & CLIP Score$\downarrow$  \\ \hline
InstPix2Pix & 23.949 & 0.813 & 0.102 & 20.087 \\
MGIE & \underline{24.994} & \underline{0.857} & \underline{0.055} & 19.528 \\ 
UltraEdit  & 21.754 & 0.744  & 0.130 & \textbf{17.891}\\
Inst-Inpaint & 22.533 & 0.710 & 0.077 & 19.581 \\
\rowcolor{gray!20}Ours & \textbf{45.742} & \textbf{0.994} & \textbf{0.004} & \underline{18.156} \\
\shline
\end{tabular}}
\vspace{-10pt}
\label{tab-remove}
\end{table}

\textbf{Instructional image inpainting.} To validate the superiority of our pipeline, we compare it with end-to-end image editing methods, including those designed for general editing tasks, namely InstPix2Pix \cite{instructpix2pix}, MGIE \cite{mgie}, and UltraEdit \cite{ultraedit}), and specifically trained for this task, Inst-Inpaint \cite{instinpaint}. The objective results are shown in the Tab. \ref{tab-remove}. Our method achieves the best results in PSNR, SSIM, and LPIPS, and reaches the second-best performance in CLIP score. The significant improvement in PSNR particularly demonstrates that the phased approach of first segmenting the object and then performing inpainting effectively preserves the pixel values in regions outside the removed object. The subjective results, as shown in Fig. \ref{fig-vs-remove}, highlight that our method not only accurately locates the object but also delivers the most natural removal effect. Other methods either fail to select the object to be eliminated or have obvious traces of editing.

\begin{figure}[t]
  \centering
  \includegraphics[width=\linewidth]{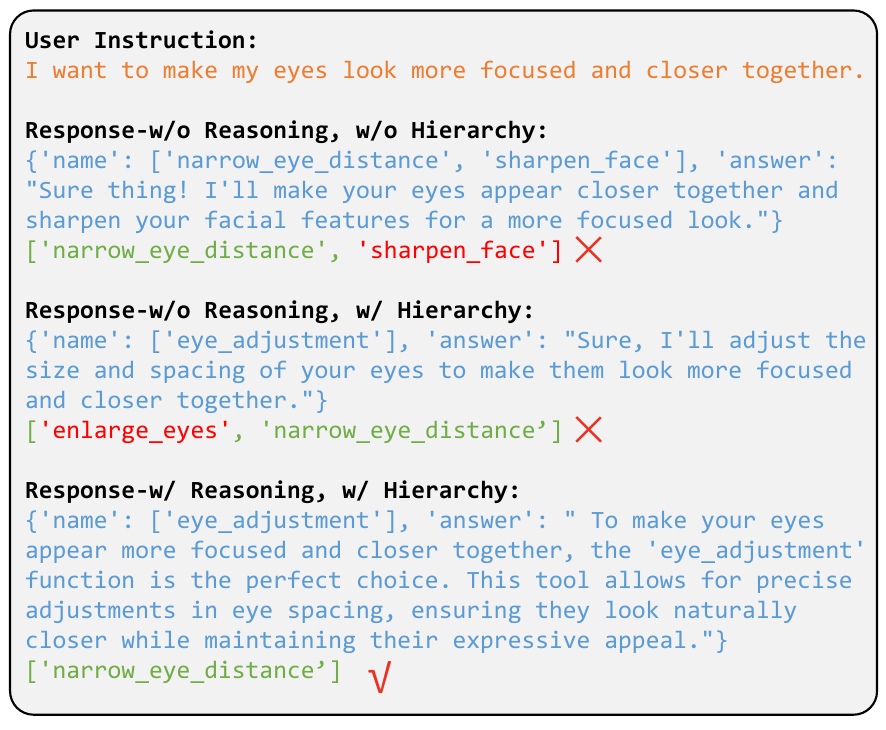}
  \vspace{-16pt}
  \caption{Comparison of an example under different configurations. Without reasoning, an irrelevant function is invoked (about the eyes, not the face). Without hierarchy, an incorrect function is invoked (about the distance between the eyes, not the size of the eyes). Only by combining both can the correct function be output.}
  \vspace{-12pt}
\label{fig-abalation} 
\end{figure}

\subsection{Ablation studies}

To demonstrate the effectiveness of our proposed design, we conduct ablation studies. The studies focused on three aspects: whether to include the phrase \textit{``You first analyze the needs of users, and then answer your reasons and suggestions for choosing these functions"} in the prompt template to guide the LLM in reasoning (``Reasoning"), whether to use a hierarchical strategy for function invocation (``Hierarchy") and how many examples in the prompt template (``Examples"). From Tab. \ref{tab-aba}, we can see that providing examples significantly improved invocation accuracy, likely due to better guidance on the LLM's output format, reducing parsing errors. However, more examples do not necessarily lead to better results; after four examples, accuracy decreased, and token consumption increased in vain. Guiding the LLM in reasoning and adopting a hierarchical strategy both improved accuracy, while the hierarchical strategy also significantly reduced token consumption. TalkPhoto ends up using the configuration labeled as (f). Taking a representative example, as shown in Fig. \ref{fig-abalation}, when reasoning is applied but not hierarchy, an irrelevant function is called. When hierarchy is applied but not reasoning, similar functions can be easily confused. Only when both are combined can accurate results be achieved.

\begin{figure*}[t]
  \centering
  \includegraphics[width=1\linewidth]{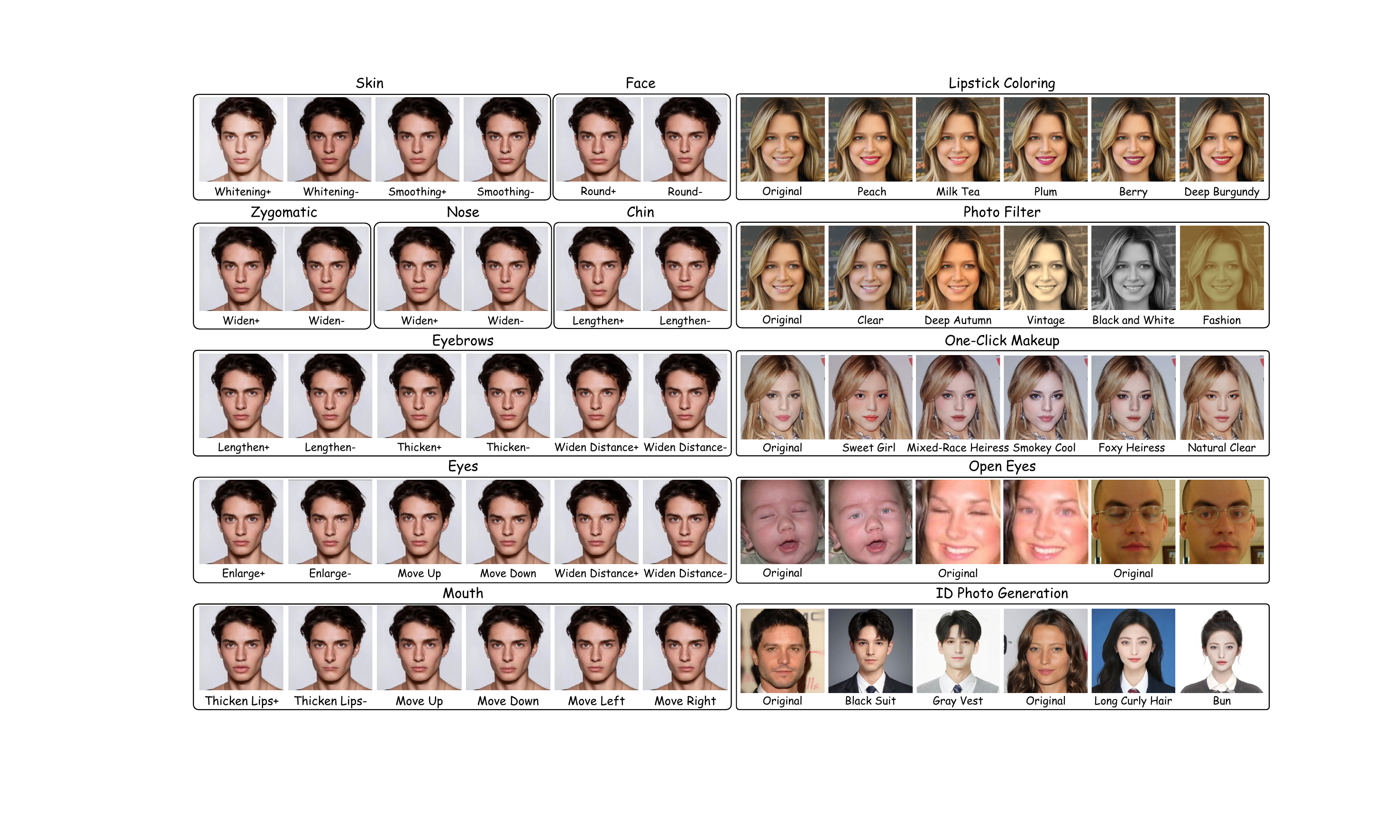}
  \vspace{-16pt}
  \caption{Visual results of \textit{Face Beautification and Shaping}, \textit{Lipstick Coloring}, \textit{Photo Filters}, \textit{One-Click Makeup}, \textit{Open Eyes}, and \textit{ID photo generation}.}
  \vspace{-10pt}
\label{fig-overall} 
\end{figure*}

\begin{figure*}[t]
  \centering
  \includegraphics[width=\linewidth]{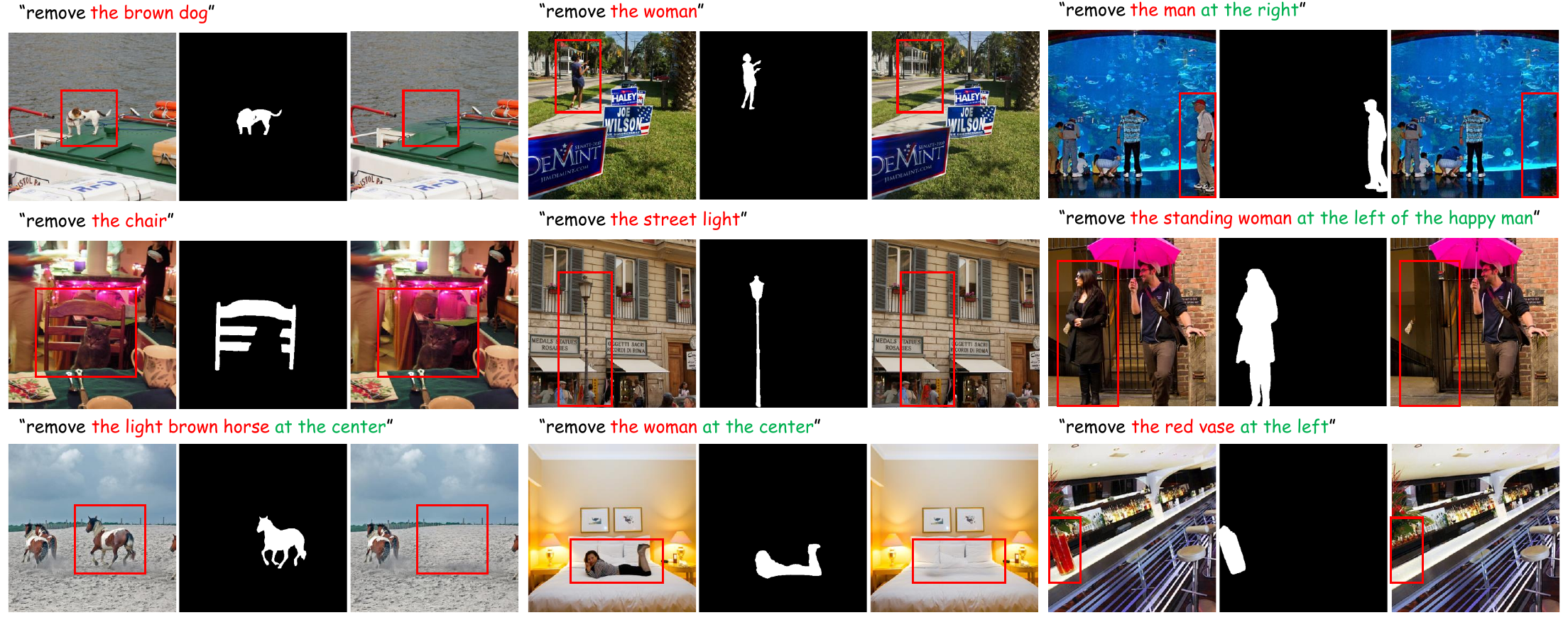}
  \vspace{-16pt}
  \caption{Visual results of \textit{Object Removal}. The columns from left to right are the original image, the segmentation mask, and our final result.}
  \vspace{-10pt}
\label{fig-add-remove} 
\end{figure*}

\subsection{Additional results}

\textbf{More Visual Results.} We provide detailed illustrations of the effects for several functions we support in Fig. \ref{fig-overall}. These include \textit{Face Beautification and Shaping}, \textit{Lipstick Coloring}, and \textit{Photo Filters} implemented using OpenCV, as well as \textit{One-Click Makeup}, \textit{ID photo generation}, and \textit{Open Eyes} on the CEW dataset\cite{cew} achieved through deep learning. This also highlights the superiority of our approach, which seamlessly integrates traditional algorithms and deep learning algorithms into a unified framework, complementing each other. It is worth mentioning that the algorithms for \textit{Face Beautification and Shaping}, \textit{Lipstick Coloring}, \textit{Photo Filters}, and \textit{Open Eyes} are entirely developed by us and not directly taken from existing algorithms. Fig. \ref{fig-add-remove} shows the effects of \textit{Object Removal} on the GQA-Inpaint dataset.

\begin{figure*}[t]
  \centering
  \includegraphics[width=1\linewidth]{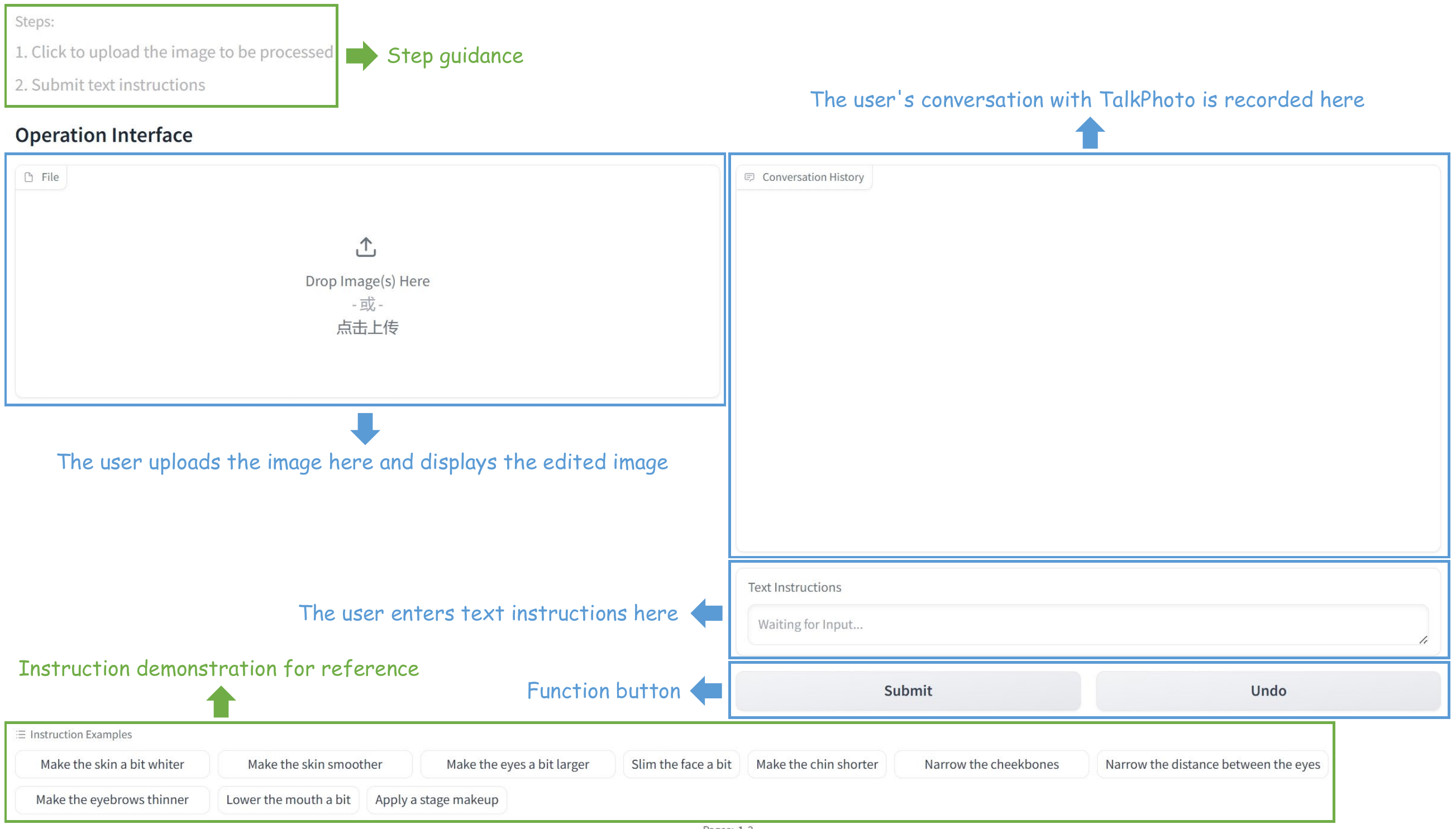}
  \caption{The graphical interface built with Gradio, mainly including an area for the user to upload and display images, an area for the user to talk to TalkPhoto, an area for the user to enter text instructions, and buttons for submit and undo. }
\label{fig-demo} 
\end{figure*}

\begin{figure*}[!h]
    \centering

    \begin{subfigure}[t]{0.45\textwidth}
        \centering
        \includegraphics[width=\linewidth]{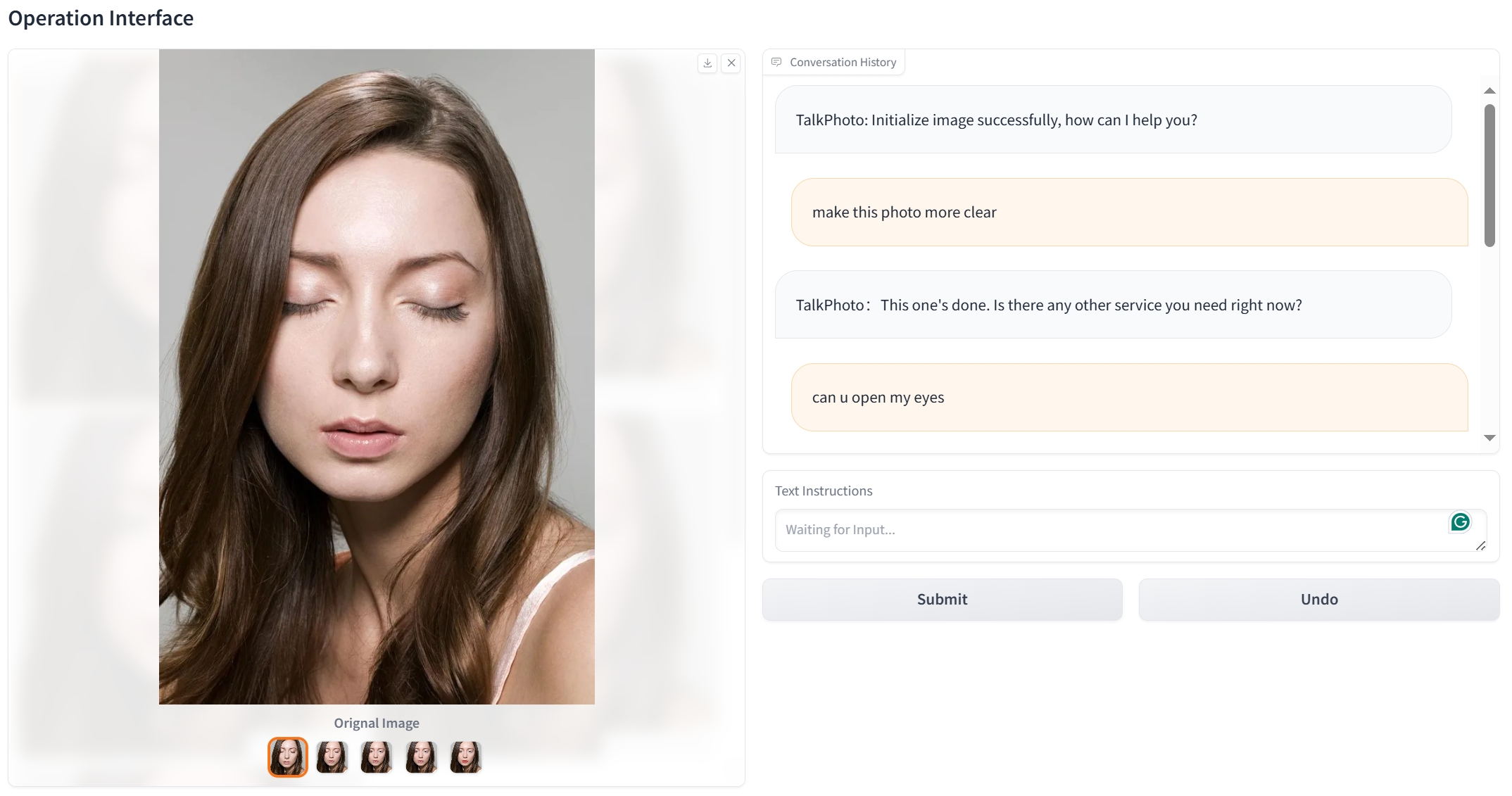}
        \caption{User uploads a photo, and TalkPhoto is waiting for the user to enter the next instruction input.}
    \end{subfigure}
    \hspace{6mm}
    \begin{subfigure}[t]{0.45\textwidth}
        \centering
        \includegraphics[width=\linewidth]{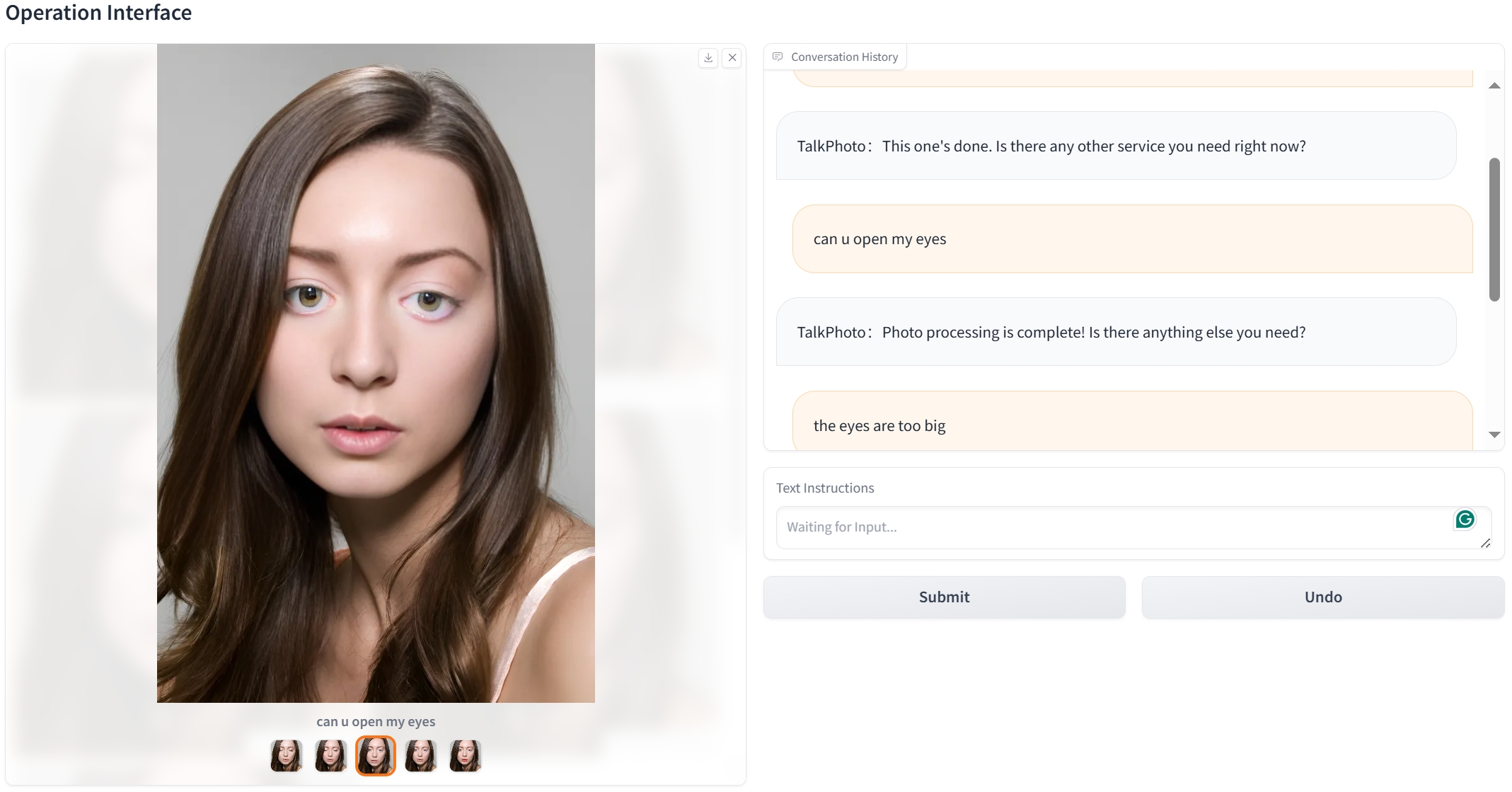}
        \caption{User inputs ``\textit{can u open my eyes}", then TalkPhoto invocates \textit{Open Eyes} and returns a text reply.}
    \end{subfigure}

    \vspace{4mm}

    \begin{subfigure}[t]{0.45\textwidth}
        \centering
        \includegraphics[width=\linewidth]{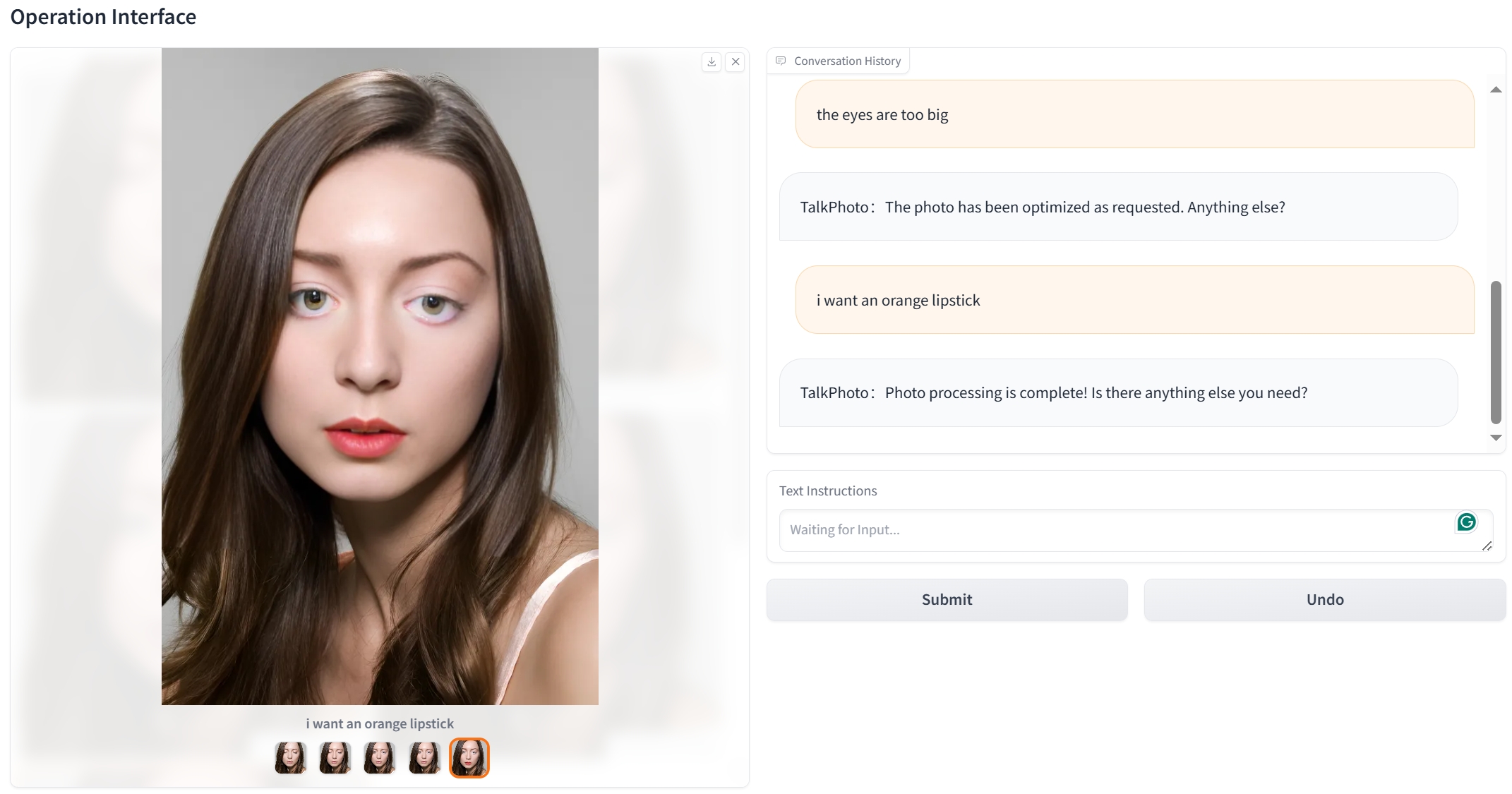}
        \caption{User inputs ``\textit{i want an orange lipstick}", then TalkPhoto invocates \textit{Lipstick Coloring}-\textit{Pure Orange} and returns a text reply.}
    \end{subfigure}
    \hspace{6mm}
    \begin{subfigure}[t]{0.45\textwidth}
        \centering
        \includegraphics[width=\linewidth]{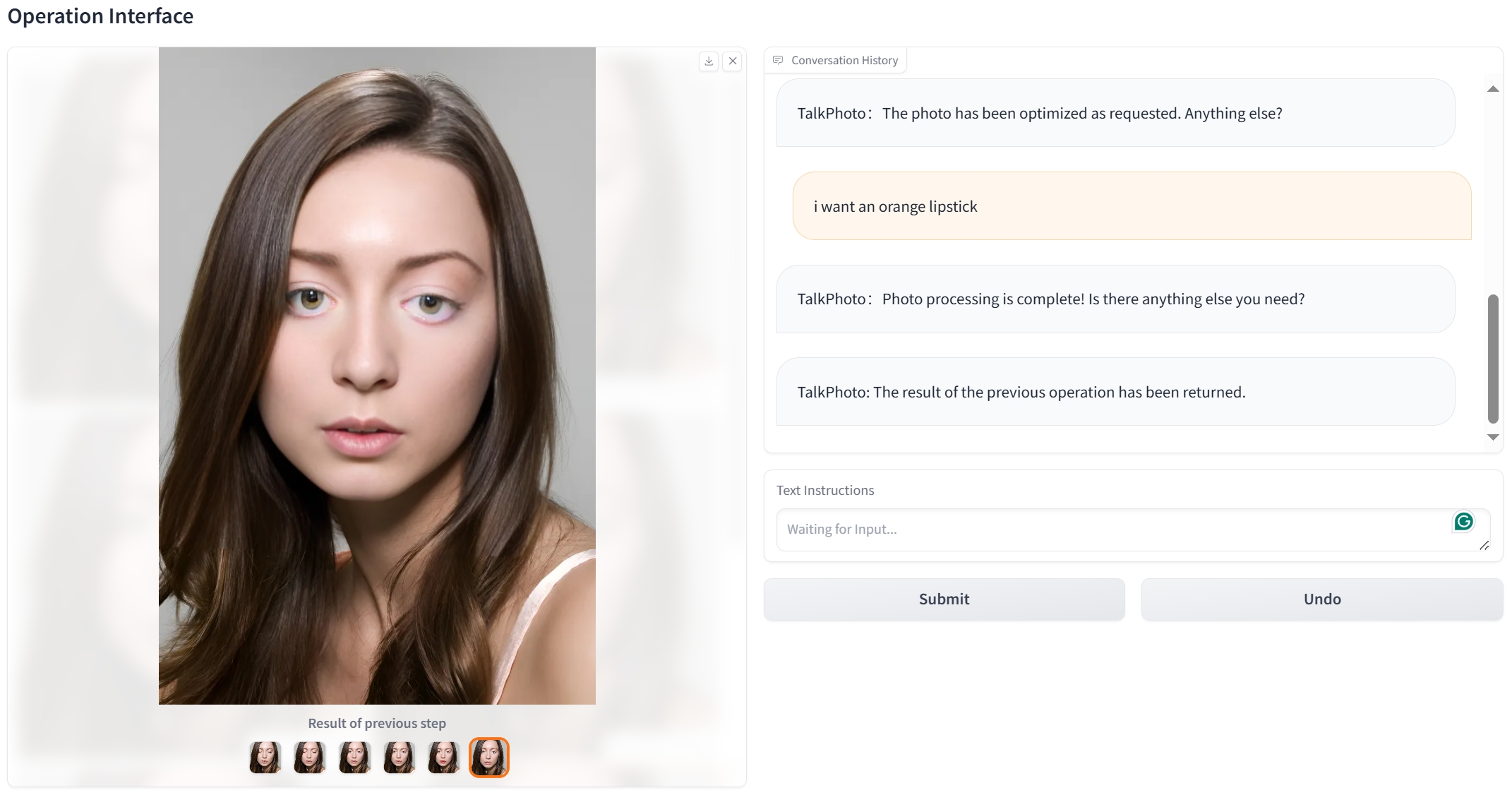}
        \caption{User clicks the ``\textit{Undo}" button, then TalkPhoto returns the previous result image and a text reply.}
    \end{subfigure}

    \caption{Demonstration of multiple rounds of conversations between users and TalkPhoto.}
    \label{fig-demo-multi}
\end{figure*}

\textbf{Local Demo.} At the same time, we deploy the LLM and other models on a local server, achieving offline functionality. The LLM used is Qwen2-7B-instruct, which can run on a single GeForce RTX 3090 (24G) GPU. Additionally, we use vLLM \cite{vllm} for further deployment of the LLM, which significantly improves loading and inference speed. As shown in Fig. \ref{fig-demo}, we use Gradio \cite{gradio} to build a graphical interface to facilitate user interaction. Fig. \ref{fig-demo-multi} demonstrates a multi-turn conversation example of TalkPhoto. The user uploads an image and inputs multiple text instructions, and TalkPhoto returns both the processed image and text responses. 

\section{Conclusion}
\label{sec5}

In this paper, we introduce TalkPhoto, a new framework for instruction-based image editing. It leverages large language models (LLMs) to comprehend and analyze user text instructions, then integrates existing image editing methods to fulfill the user's needs. We enhance the flexibility of function invocation through a carefully designed prompt template. We improve invocation accuracy and reduce token consumption per edit by simplifying functions and introducing a hierarchical strategy. Additionally, TalkPhoto supports combining existing models to handle complex image editing tasks, such as instructional image inpainting, to achieve better object removal results. Massive experiments show that TalkPhoto's function invocation capability significantly outperforms existing LLM-based tool usage frameworks and delivers strong performance across various tasks. This framework provides an easier way for ordinary users to edit images, without the need to learn any specialized knowledge or spend time searching for the latest techniques for each task. Our work also demonstrates the great potential of combining pre-trained large language models with large visual models, which will be beneficial for future vision-and-language research. 

\textbf{Discussion.} Current tool-augmented language models primarily depend on the performance of LLMs, and we are no exception. Our method can also be seen as a summary of practical insights on how to better utilize LLMs to accomplish specific tasks during experimentation.
It does not require pre-training LLMs with JSON formats. Compared to pre-packaged prompt engineering, our approach offers a more flexible and efficient way to use various tools through LLMs. In the future, We will consider integrating reinforcement learning, using user feedback to adjust the model output, thereby enhancing editing stability and effectiveness. We will also introduce multimodal large language models (MLLMs) to improve the combined understanding of images and text, thereby automatically recommending appropriate editing operations to users.

\bibliographystyle{CVMbib}
\bibliography{refs}

\vfill

\end{document}